\documentclass[conference]{IEEEtran}
\IEEEoverridecommandlockouts
\renewcommand{\figurename}{Figure}
\usepackage{cite}
\usepackage{amsmath,amssymb,amsfonts}
\usepackage{algorithmic}
\usepackage{graphicx}
\usepackage{textcomp}
\usepackage{xcolor}
\usepackage{url}
\usepackage{amssymb}
\usepackage{makecell}  
\usepackage{tabularx}

\usepackage[capitalize]{cleveref}
\crefname{equation}{Eq.}{Eqs.}

\renewcommand{\arraystretch}{1.5}

\def\BibTeX{{\rm B\kern-.05em{\sc i\kern-.025em b}\kern-.08em
    T\kern-.1667em\lower.7ex\hbox{E}\kern-.125emX}}

\begin{document}

\title{VAST: V2X/Dynamic Map-Aware Autonomous Driving Systems Validation Toolchain}

\author{
    \IEEEauthorblockN{Shunsuke Ito}
    \IEEEauthorblockA{
        Graduate School of 
        Science and Engineering \\
        Saitama University}
    \and
    \IEEEauthorblockN{Takuya Azumi}
    \IEEEauthorblockA{
        Academic Association 
        (Graduate School of Science and Engineering) \\
        Saitama University
    }
}

\maketitle

\begin{abstract}
Cooperative autonomous driving in the IoT-to-Edge-to-Cloud continuum requires system-level validation across vehicles, infrastructure sensors, edge-side Dynamic Map services, and in-vehicle autonomous-driving stacks. This paper presents VAST, a V2X/Dynamic Map-aware validation toolchain that connects Scenic, Scenario Simulator v2, AWSIM, Autoware, and SIM-LDM. VAST does not introduce a new search algorithm; instead, it addresses interoperability challenges, including Lanelet2-to-Scenic mapping, ROS~2-based co-simulation through SS2, Dynamic Map object injection into Autoware, and collection of TTC, PET, collision, timeout, and performance measurements. In occluded-intersection scenarios, Lanelet2-compatible constrained sampling increases the edge-case discovery rate from 40.0\% to 80.0\% and reduces the average time per discovered edge case from 259.7~s to 110.4~s. Under the same generated scenario distribution, Dynamic Map availability reduces the collision rate from 78.0\% to 40.0\% and increases non-collision outcomes from 22.0\% to 60.0\%, with statistically significant TTC/PET shifts. A throughput study with 1--16 NPCs shows that sampling remains below 0.1~s, whereas AWSIM/Autoware execution and restart overhead dominate runtime. These results position VAST as a practical validation infrastructure for cooperative autonomous-driving CPSs.
\end{abstract}


\section{Introduction}\label{sec:intro}
In the IoT-to-Edge-to-Cloud continuum, cooperative autonomous driving is emerging as a cyber-physical system that connects vehicles, roadside infrastructure sensors, edge-side Dynamic Map services, and in-vehicle autonomous-driving stacks through Vehicle-to-Everything (V2X) communication~\cite{V2X_survey,V2X_example}. By using infrastructure-side observations, such systems can extend vehicle situational awareness beyond onboard sensor range and line of sight~\cite{cyo-IV}. Therefore, validation must evaluate the complete information path from infrastructure sensing and Dynamic Map management to vehicle-side perception, planning, and control, rather than isolated onboard sensing or scenario generation alone.

Within this information path, Dynamic Map~\cite{SIPAdus-DynamicMap} plays a central role by aggregating infrastructure-sensor and traffic-signal information and sharing it with vehicles through V2X communication~\cite{iLDM,liveMap}. However, evaluating Dynamic Map-based cooperative driving in real environments remains costly and difficult to control, because infrastructure sensing, edge-side data management, V2X delivery, and vehicle behavior vary simultaneously~\cite{offload}. Simulation-based validation is therefore essential, but it must reproduce not only traffic scenes but also the Dynamic Map data path and its interface to vehicle-side planning/control software~\cite{Riedmaier_SBT_Survey}.

Scenario-based testing is widely used to evaluate autonomous-driving systems under controlled and repeatable conditions, especially for rare but safety-critical edge cases~\cite{Riedmaier_SBT_Survey}. 
However, existing scenario-generation and simulation tools do not sufficiently support reproducible evaluation of V2X/Dynamic Map effects in Autoware/AWSIM environments. 
In particular, a validation loop that connects probabilistic scenario specification, Lanelet2 map geometry, Dynamic Map injection, and vehicle-side planning/control execution remains insufficiently established.

Based on the challenges described above, this research addresses the following Research Questions (RQs):\\
    \textbf{RQ1}: Can a Lanelet2-compatible generation pipeline efficiently produce reproducible edge cases in an Autoware/AWSIM/Dynamic Map validation environment?\\
    \textbf{RQ2}: How does Dynamic Map availability affect the system-level behavior of Autoware?\\
    \textbf{RQ3}: What limits the throughput of an integrated V2X/Dynamic Map-aware validation pipeline?

To answer these RQs, this paper proposes VAST, a V2X/Dynamic Map-aware validation toolchain integrating Scenic~\cite{Scenic}, Scenario Simulator v2, AWSIM, Autoware, and SIM-LDM to generate, execute, and evaluate edge-case scenarios for system-level cooperative-driving validation.

The contributions of this paper are summarized as follows:

\begin{itemize}
    \item \textbf{Lanelet2-compatible V2X/Dynamic Map-aware validation toolchain.}
    VAST is a ROS~2-based validation toolchain that integrates Scenic, Scenario Simulator v2, AWSIM, Autoware, and SIM-LDM.
    Rather than proposing a new search algorithm, the technical novelty of VAST lies in the Lanelet2-to-Scenic mapping layer, the step-locked Scenic/SS2 co-simulation mechanism, and the Dynamic Map injection path for paired Autoware evaluation.

    \item \textbf{Controlled evaluation of Dynamic Map availability.}
    VAST includes a paired execution pipeline that evaluates the same generated scenario distribution with and without V2X-derived Dynamic Map information. 
    This enables quantitative analysis of how Dynamic Map availability affects the system-level behavior of Autoware using collision, timeout, TTC, and PET metrics.

    \item \textbf{Throughput characterization of the integrated validation pipeline.}
    The execution cost of the Scenic/SS2/AWSIM/Autoware/SIM-LDM pipeline is evaluated with 1--16 NPCs. 
    The analysis identifies practical throughput bottlenecks, showing that AWSIM/Autoware execution and restart overhead dominate runtime rather than Scenic sampling.
\end{itemize}

    The remainder of this paper is organized as follows.
    Section~\ref{sec:SystemModel} presents an overview of the foundational tools, Section~\ref{sec:Design} describes the proposed method, Section~\ref{sec:Evaluation} evaluates the proposed method, Section~\ref{sec:Related} reviews related work, and Section~\ref{sec:Conclusion} concludes the paper.

\section{System Model}\label{sec:SystemModel}
    
\begin{figure*}[t]
        \centering
        \includegraphics[width=\textwidth]{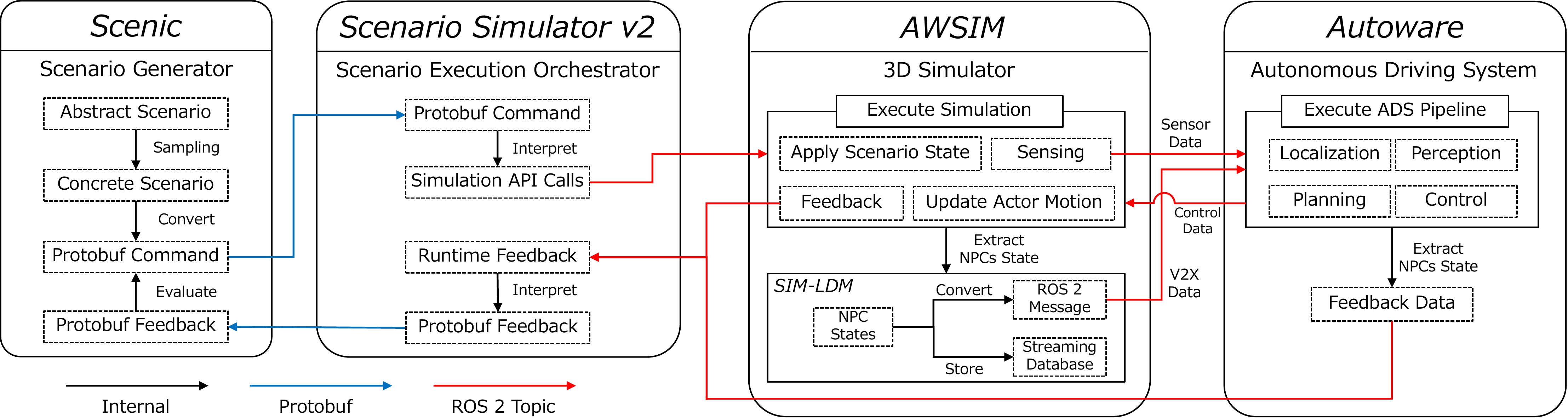}
        \caption{Integrated system model and validation workflow.}
        \label{fig:SystemModel}
        \vspace{-4mm}
    \end{figure*}

    The proposed toolchain builds on five foundational technologies and tools.
    The proposed system integrates Scenic, a probabilistic scenario description language, Dynamic Map, SIM-LDM for reproducing Dynamic Map on the simulator AWSIM, Autoware autonomous driving software, and Scenario Simulator, a scenario execution framework.
    An overview of the system architecture is illustrated in Figure~\ref{fig:SystemModel}.
    In this model, roadside sensors and AWSIM correspond to cyber-physical data sources, SIM-LDM and DSMS form the edge-side information layer, and Autoware represents the in-vehicle compute node evaluated by VAST.

    \subsection{Scenic}\label{subsec:Scenic}
        Scenic is a domain-specific probabilistic scenario description language designed for scenario specification~\cite{Scenic}.
        A scenario is defined as a probability distribution over scenes, which are configurations of physical objects and agents, and is characterized by abstract specification using probability distributions and constraints.
        This probabilistic representation enables the generation of diverse concrete scenes through sampling from a single scenario specification, thereby facilitating efficient edge-case exploration.
        
        Scenic is adopted in this research because it is the most suitable mature scenario description language with probabilistic scenario generation capabilities.
        OpenSCENARIO~\cite{OpenSCENARIO_XML,OpenSCENARIO_DSL}, the industry standard, has a 1.x series specialized for deterministic scenario description (Concrete Scenario), while the 2.0 series introduced constraint-based description enabling Logical Scenario specification.
        Compared to probabilistic programming languages such as Scenic, however, OpenSCENARIO differs in its approach to intuitive scene generation and sampling expressiveness using probability distributions.
        Scenic is designed with probability distributions and constraints as core language features, natively supporting two types of constraints: hard constraints and soft constraints.
        Combining these constraints enables flexible spatial constraint specifications, such as placing pedestrians in regions intersecting the ego vehicle's travel path with high probability, or placing NPC vehicles within a specified distance range from the ego.
        Furthermore, because Scenic is Python-based, Scenic offers superior readability and ease of learning compared to OpenSCENARIO DSL or XML, and integration with existing Python libraries is straightforward.
        The language also supports temporal logic specifications for dynamic scenarios, enabling the description of scenarios involving temporal changes.
    
    \subsection{Dynamic Map}\label{subsec:DynamicMap}
        Dynamic Map is a platform based on the Local Dynamic Map (LDM) defined by ETSI~\cite{ETSI_LDM} that aggregates and shares information from vehicles and infrastructure sensors via V2X communication~\cite{SIPAdus-DynamicMap}.
        The integrated system model in Figure~\ref{fig:SystemModel} places Dynamic Map in the edge-side information layer that bridges infrastructure sensing and vehicle-side planning.
        Dynamic Map manages heterogeneous road-environment information as layered data according to update frequency and information type, ranging from static map information to dynamic traffic information.
        
        Dynamic Map adopts stream processing and in-memory management to process continuously incoming data.
        A Data Stream Management System (DSMS)~\cite{DSMS_Survey} performs these tasks, enabling immediate processing upon data arrival and low-latency data access.
        Geographic distribution of edge nodes enables local processing of information within defined areas, including object detection from sensor data~\cite{RAGE_nogu,CCNC_nogu}, while cloud nodes aggregate data from broader regions.

    \subsection{AWSIM and Autoware}\label{subsec:AWSIM_Autoware}
        AWSIM~\cite{AWSIM} is a Unity-based open-source autonomous driving simulator that integrates seamlessly with Autoware~\cite{Autoware} through ROS~2-native communication.
        Autoware provides autonomous driving functionalities, such as Perception, Planning, and Control, through a modular architecture, and has a track record of deployment on real vehicles.
        The proposed method adopts a configuration in which Autoware controls the ego vehicle on AWSIM.
        
        Dynamic Map integration into the simulation environment is achieved through SIM-LDM~\cite{SIM-LDM}, a framework that extends AWSIM to generate Dynamic Map data.
        SIM-LDM replaces real-world data with simulator-generated data, enabling research and development utilizing Dynamic Map without requiring real environments.
        A ROS~2 node implemented within Autoware requests data from the Data Stream Management System of SIM-LDM and obtains V2X information for use in autonomous driving control.
    
    \subsection{Scenario Simulator (SS2)}\label{subsec:ScenarioSimulator}      
        Scenario Simulator v2 (hereafter referred to as SS2) is a test framework for executing scenario-based testing targeting Autoware, comprising a three-layer structure of a test management layer, a scenario interpretation layer, and a simulation execution layer~\cite{SS}.
        Integration with AWSIM and Autoware has already been achieved, enabling flexible extension through clear API boundaries.
        The supported scenario formats, however, are limited to OpenSCENARIO 1.2~\cite{OpenSCENARIO_XML}, and Scenic is not supported.
        The proposed method extends the scenario interpretation layer to realize Scenic support.

\section{Design \& Implementation}\label{sec:Design}
    The central contribution of the proposed framework is a V2X/Dynamic Map-aware validation toolchain that makes generated scenarios executable in an Autoware/AWSIM environment and comparable across Dynamic Map availability conditions.
    The framework integrates the probabilistic constraint engine of Scenic as one input component, but the validation capability is realized by the surrounding co-simulation, Dynamic Map injection, paired execution, and metric collection pipeline.
    This integration is non-trivial because Scenic, SS2, AWSIM, SIM-LDM, and Autoware use different map abstractions, execution models, and data interfaces, which must be synchronized without changing the generated scenario itself.
    
    The overall architecture in Figure~\ref{fig:SystemModel} consists of a Scenic-driven co-simulation loop and a V2X-enabled data path.
    In the co-simulation loop, Scenic acts as the time master and exchanges Protobuf messages with SS2 over ZeroMQ, while SS2 controls AWSIM and Autoware through ROS~2 interfaces.
    The middleware boundary preserves ROS~2 topics inside SS2/AWSIM/Autoware and uses ZeroMQ/Protobuf only at the Scenic interface, keeping probabilistic scenario generation separate from the ROS~2 execution graph.
    In the V2X-enabled data path, AWSIM streams object information to SIM-LDM, and Autoware consumes the resulting Dynamic Map information as perception input.
    The design first describes how Scenic supplies controllable stress inputs, then details the Scenic-driven co-simulation architecture, and finally presents the V2X-enabled quantitative evaluation pipeline.

    \subsection{Scenario Generation as a Validation Input}\label{subsec:ConstraintBasedGeneration}
        Scenario generation in VAST is realized through a closed loop comprising three phases: scene sampling, simulation execution, and observation feedback.
        First, a concrete scene consisting of the initial position, velocity, and heading of each entity is automatically sampled according to the probability distributions and spatial constraints specified in the Scenic scenario definition.
        The generated scene is then reflected in SS2 and AWSIM, and a real-time simulation is executed in conjunction with the Autoware autonomous driving software stack.
        Finally, the observations from each simulation run, including entity states, TTC, PET, and collision status, are returned to Scenic and used for termination-condition evaluation and success determination.

        The spatial constraint functionality of Scenic is the core of this loop.
        The two constraint types and representative usage examples are summarized in Table~\ref{tab:scenic_constraints}.
        Hard constraints reject samples that violate the specified condition, whereas soft constraints require the condition to hold with at least a given probability.
        Combining these constraint types generates, with high probability, scenes in which an NPC is inside an intersection and within 15~m of the ego vehicle, enabling VAST to supply stress inputs that are rare under random sampling.

        \begin{table}[t]
            \centering
            \caption{Scenic Constraint Types and Usage Examples}
            \label{tab:scenic_constraints}
            {\renewcommand{\arraystretch}{1.3}
            \begin{tabularx}{\columnwidth}{|l|X|} \hline
                \textbf{Type} & \textbf{Syntax / Example} \\ \hline
                Hard constraint &
                    \texttt{require <condition>} \newline
                    e.g., \texttt{require npc in intersection} \\ \hline
                Soft constraint &
                    \texttt{require[p] <condition>} \newline
                    e.g., \texttt{require[0.8] distance to npc < 15} \\ \hline
            \end{tabularx}}
            \vspace{-6mm}
        \end{table}

        The incompatibility of map formats renders the constraint functionality of Scenic inoperative without modification, as Scenic natively supports only the OpenDRIVE map format while AWSIM and Autoware adopt the Lanelet2 format.
        To resolve this incompatibility, an Environment Module is implemented that parses Lanelet2 map data in OSM-XML format, constructs lane polygons from the left and right boundary point sequences of each lanelet, and exposes them as \textit{PolygonalRegion} objects conforming to the Scenic standard.
        Specifically, the module generates five region types: \textit{road} (all drivable area), \textit{intersection} (area within intersections), \textit{straight\_road} (non-intersection roads), \textit{sidewalk}, and \textit{crosswalk}, and additionally provides a \textit{roadDirection} vector field derived from lanelet connectivity to Scenic.
        Scenario authors can therefore write map-based constraints in standard Scenic syntax, while the Environment Module internally evaluates sampling feasibility using the Lanelet2-derived region geometry.
        During sampling, inter-object collision detection and road-boundary interference checking are performed automatically, and physically inconsistent scenes are rejected prior to simulation execution.

    \subsection{Scenic-Driven Co-Simulation Architecture}\label{subsec:ScenarioSimulatorExtension}
        \begin{figure}[t]
            \centering
            \includegraphics[width=\linewidth]{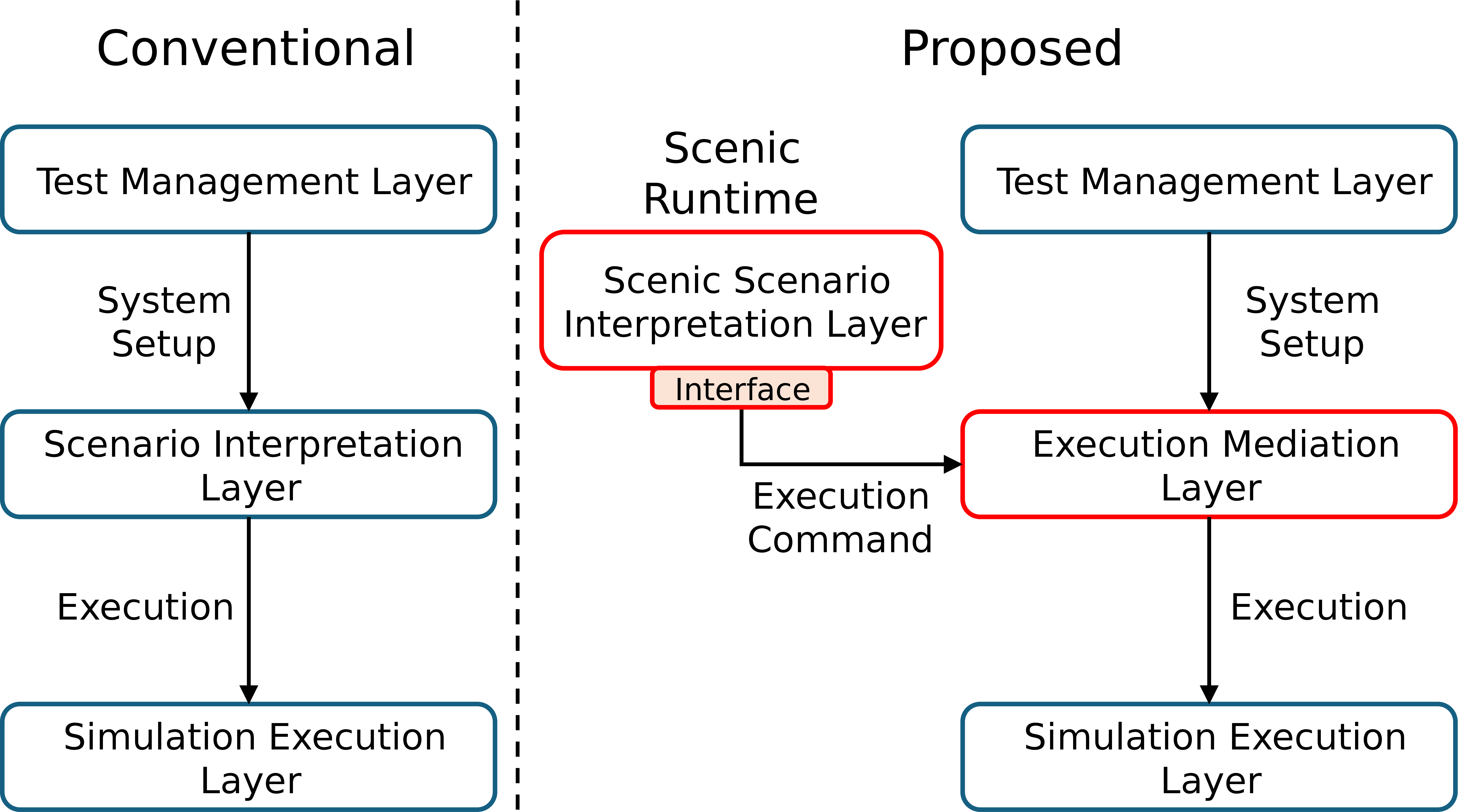}
            \caption{Comparison of the conventional SS2 (left) and the proposed extension (right). A Scenic Interpreter is added as a new ROS~2 lifecycle node alongside the existing OpenSCENARIO Interpreter, connected to Scenic via ZeroMQ REQ-REP.}
            \label{fig:Diff_ScenarioSimulator}
            \vspace{-4mm}
        \end{figure}
        Realizing the closed loop described above requires integrating Scenic, which operates probabilistically and in a step-driven manner, with the existing SS2, which was designed as a deterministic OpenSCENARIO execution engine.
        Two challenges arise in this integration: the transfer of control authority between Scenic and SS2 that operate asynchronously, and the design of time advancement that accounts for the non-deterministic state transitions of Autoware.

        As shown in Figure~\ref{fig:Diff_ScenarioSimulator}, the scenario interpretation layer of SS2 is extended by adding a Scenic-dedicated interpreter as a new ROS~2 lifecycle node that operates in parallel with the existing OpenSCENARIO interpreter.
        ZeroMQ REQ-REP is adopted for communication between the Scenic runtime and SS2, and messages are serialized using Protobuf.
        The Protobuf schema defines entity spawn requests (\textit{SpawnEgoRequest}, \textit{SpawnNPCVehicleRequest}, and \textit{SpawnPedestrianRequest}) and the message pair that forms the core of the simulation loop (\textit{UpdateNPCStatesRequest} and \textit{UpdateNPCStatesResponse}).
        At each step, Scenic evaluates the behavior of each entity and sends actions such as \textit{FollowLaneAction} (route following along specified lanelets), \textit{FollowTrajectoryAction} (trajectory following along a waypoint sequence), and \textit{SetVelocityAction} to SS2.
        SS2 executes the actions, then calls \texttt{api\_->updateFrame()} to advance the simulation clock by one step, and returns the states of all entities to Scenic.

        A key design decision is that \texttt{api\_->updateFrame()} is called exclusively upon receipt of an \textit{UpdateNPCStatesRequest} from Scenic, making Scenic the sole time master of the simulation.
        This step-lock mechanism ensures that AWSIM never advances its clock until Scenic issues a request, guaranteeing deterministic causal ordering between the behavior logic of Scenic and the physics simulation of AWSIM.
        Control of the ego vehicle is delegated entirely to Autoware; Scenic specifies only the initial position and goal position of the ego vehicle, and all decisions regarding path planning, speed control, and obstacle avoidance are handled by Autoware, faithfully reproducing the behavior of an actual autonomous driving system.

    \subsection{V2X-Enabled Evaluation Pipeline}\label{subsec:DynamicMapIntegration}
        Quantitative evaluation of the collision avoidance effect of Dynamic Map (RQ2) requires an environment in which the same automatically generated edge-case scenario can be executed with and without V2X information.
        Realizing this requires constructing a pipeline that injects entity information from AWSIM into Dynamic Map in real time and enables Autoware to consume the injected data as perception input.

        Ground truth information (position and velocity) of NPC vehicles and pedestrians in AWSIM is streamed to the Data Stream Management System (DSMS) of SIM-LDM via socket communication and Apache Kafka.
        A V2X interface node implemented within Autoware queries DSMS for object information on the current driving lane and inside intersections, and delivers the retrieved object list to the object recognition module of Autoware (\textit{/perception/object\_recognition}).
        Recognition results from onboard LiDAR and cameras are fused with V2X-derived object information inside the recognition module, and the integrated occupancy grid is provided to the path planning module of Autoware.
        V2X availability is toggled by a startup flag of the interface node; executing the same scenario under both conditions and comparing TTC and PET distributions as well as collision occurrence rates enables quantitative evaluation of the contribution of Dynamic Map.
        The current implementation focuses on isolating the behavioral effect of Dynamic Map availability and therefore uses simulator-provided object states as the Dynamic Map input. Network delay, packet loss, jitter, stale updates, and infrastructure-sensor uncertainty are not modeled in the present experiments; these factors are treated as robustness dimensions for future extensions.

\section{Evaluation}\label{sec:Evaluation}
    The evaluation answers RQ1--RQ3 defined in Section~\ref{sec:intro}.
    The setup is described first, followed by the methodology and results for each RQ.
    
    \subsection{Experimental Setup}\label{subsec:ExperimentalSetup}
        The experiments are conducted in an environment that integrates all components of VAST.
        The experimental environment is presented in Table~\ref{tab:experimental_environment}.
        \begin{table}[!t]
            \centering
            \caption{Experimental Environment}
            \label{tab:experimental_environment}
            \renewcommand{\arraystretch}{1.2}
            \begin{tabularx}{\columnwidth}{|l|l|X|}
                \hline
                \textbf{Category} & \textbf{Item} & \textbf{Specification} \\
                \hline
                Hardware & CPU & Intel Core i7-13700KF \\
                \cline{2-3}
                & GPU & GeForce RTX 4090 (24 GB) \\
                \cline{2-3}
                & RAM & 128 GB \\
                \hline
                Software & Scenic & 3.0 \\
                \cline{2-3}
                & AWSIM & 2.0.0 \\
                \cline{2-3}
                & Autoware Universe & 0.48.0 \\
                \cline{2-3}
                & Scenario Simulator v2 & 18.2.1 \\
                \hline
            \end{tabularx}
            \vspace{-4mm}
        \end{table}
        
        The evaluation map is a Lanelet2-format urban intersection map with buildings that create blind corners.
        The baseline scenario places an NPC vehicle or pedestrian on an intersecting approach as the ego vehicle enters the intersection, reproducing cases that are difficult to detect using onboard sensors alone.
        
        Post-Encroachment Time (PET) and Time-to-Collision (TTC), employed in prior research~\cite{Detection_CornerCase}, are adopted for edge-case determination.
        PET represents the time difference when two objects pass through the same point at an intersection, with smaller values indicating higher proximity.
        TTC represents the predicted time until collision if two objects maintain their current speed and direction.
        Following prior research~\cite{Detection_CornerCase}, PET $<$ 2.0 seconds and TTC $<$ 3.0 seconds are used as edge-case thresholds.
    
    \subsection{RQ1: Stress-Input Reproducibility}\label{subsec:RQ1}
        RQ1 compares an unconstrained Baseline condition and a constrained Proposed condition on the intersection blind-corner scenario.
        The Baseline condition randomly placed NPC vehicles over the entire approach lane and released them after a random delay, whereas the Proposed condition restricted placement to a band near the intersection and released NPC vehicles when the ego vehicle approached the conflict area.
        Under each condition, 25 scenarios were generated and executed, and the minimum PET and minimum TTC of each run were measured for edge-case determination.
        One run under the Proposed condition initially failed because of an infrastructure error; however, a rerun restored 25 valid runs.

\begin{figure}[t]
            \centering
            \includegraphics[width=0.88\linewidth]{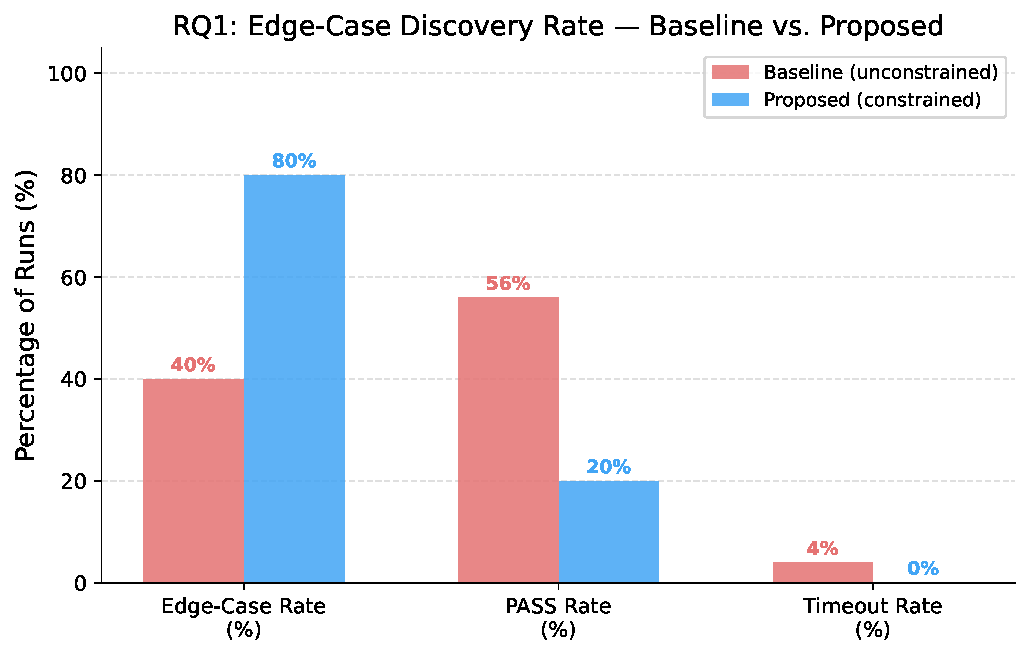}
            \caption{Comparison of edge-case, pass, and timeout rates in RQ1.}
            \label{fig:rq1_edge_case_rate}
            \vspace{-4mm}
        \end{figure}

        Spatial constraints substantially improved the stress-input yield for the downstream validation pipeline.
        As shown in Figure~\ref{fig:rq1_edge_case_rate}, the Proposed condition achieved an edge-case discovery rate of 80.0\% (20/25), whereas the Baseline condition achieved 40.0\% (10/25).
        The pass rate decreased from 56.0\% to 20.0\%, and the single timeout observed in the Baseline condition disappeared in the Proposed condition.
        These rates indicate that spatial constraints concentrated sampling on cases in which the ego vehicle and an NPC vehicle could intersect within the conflict area, increasing the number of useful validation runs per execution budget.

        Spatial constraints also changed the stress-input composition.
        All 20 edge cases under the Proposed condition were PET threshold breaches, indicating consistent generation of close crossing timing.
        The Baseline condition produced four PET threshold breaches, four TTC threshold breaches, and two near-miss events, defined as simultaneous PET and TTC threshold violations.
        This composition suggests that unconstrained generation discovered hazardous situations incidentally, whereas constrained generation repeatedly produced short-interval intersection crossings.

\begin{figure}[t]
            \centering
            \includegraphics[width=0.88\linewidth]{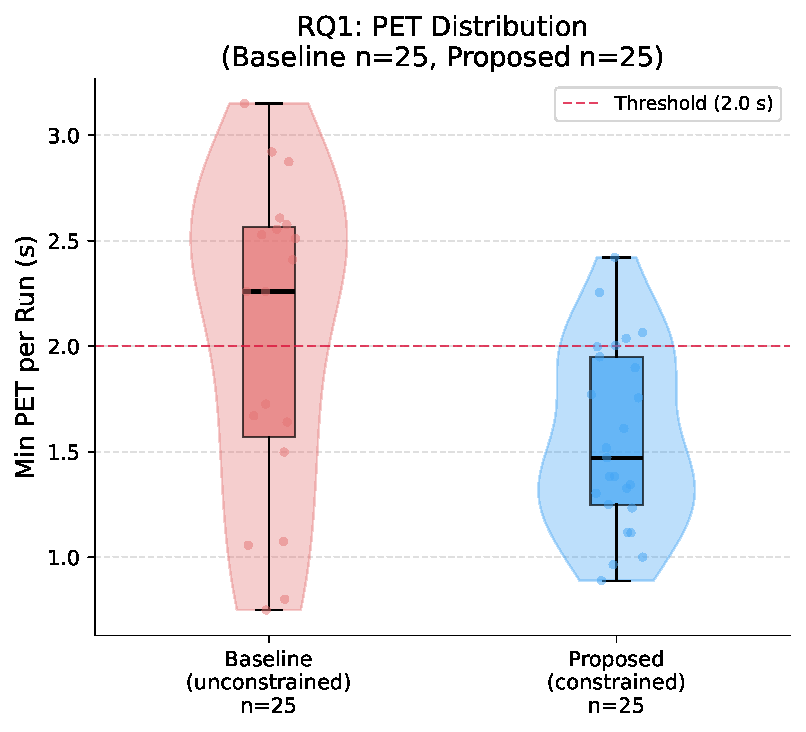}
            \caption{Distribution of the minimum PET per run in RQ1. Each point indicates one run, and the dashed line denotes the PET threshold of 2.0\,s used for edge-case detection.}
            \label{fig:rq1_pet_distribution}
            \vspace{-4mm}
        \end{figure}

        Spatial constraints shifted the PET distribution toward more critical values.
        As shown in Figure~\ref{fig:rq1_pet_distribution}, the mean minimum PET under the Proposed condition was 1.56\,s and the median was 1.47\,s, both of which were below the 2.0\,s edge-case threshold.
        Under the Baseline condition, the mean minimum PET was 2.05\,s and the median was 2.26\,s for the 19 runs in which PET was observed.
        The standard deviation decreased from 0.73\,s to 0.42\,s, indicating tighter control of crossing timing.
        The six Baseline runs without PET measurements corresponded to cases in which the NPC vehicle did not traverse the intersection sufficiently.

\begin{figure}[t]
            \centering
            \includegraphics[width=0.88\linewidth]{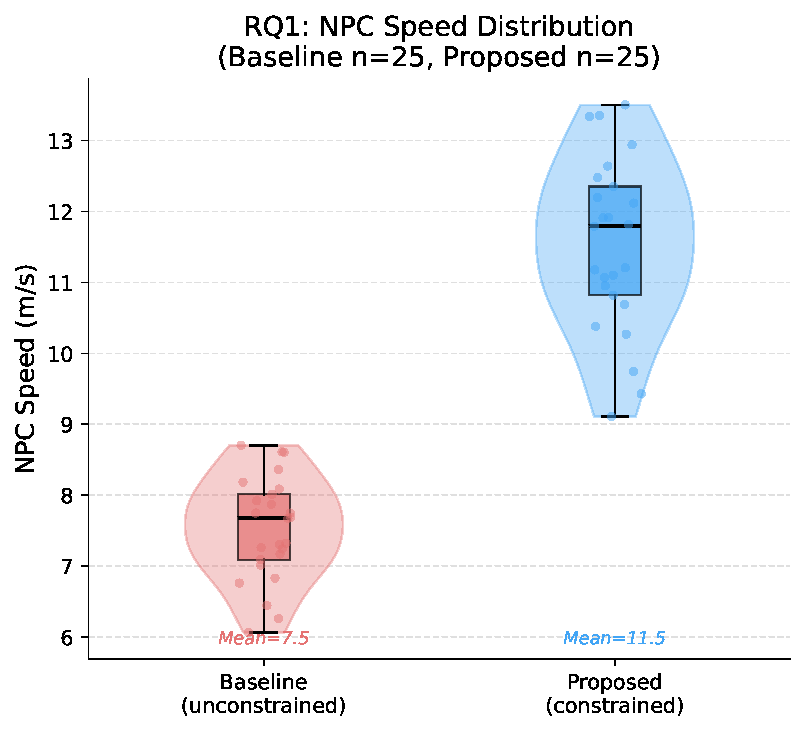}
            \caption{Distribution of NPC speed in RQ1. The constrained condition sampled higher approach speeds than the unconstrained baseline.}
            \label{fig:rq1_npc_speed_distribution}
            \vspace{-4mm}
        \end{figure}

        The higher density of hazardous scenarios was associated with NPC speed and release timing.
        As shown in Figure~\ref{fig:rq1_npc_speed_distribution}, the mean NPC speed was 11.53\,m/s under the Proposed condition, compared with 7.52\,m/s under the Baseline condition.
        The Proposed condition also controlled the NPC release distance within 20.21--34.07\,m, producing high-speed approaches near the intersection and reducing the temporal gap between ego-vehicle entry and NPC arrival.

        Validation-input efficiency also improved.
        The mean execution time per scenario decreased from 103.9\,s to 88.3\,s, and the mean time required to discover one edge case decreased from 259.7\,s to 110.4\,s, corresponding to approximately 2.4$\times$ higher efficiency.
        These results answer RQ1 by showing that Scenic-based spatial constraints doubled the edge-case discovery rate and reduced the time required to obtain one stress input to less than half of the unconstrained baseline.

    \subsection{RQ2: Collision Avoidance with Dynamic Map}\label{subsec:RQ2}
        RQ2 evaluates collision avoidance in a low-visibility intersection where a signal-ignoring vehicle approaches from a crossing direction.
        The Baseline condition uses only onboard LiDAR and camera sensing, whereas the Proposed condition adds V2X information via Dynamic Map as described in Section~\ref{subsec:DynamicMapIntegration}.
        Because Autoware planning and control include non-deterministic behavior, 50 runs were conducted under each condition using scenarios sampled from the same scenario distribution with matched random seeds.
        The evaluation compares the non-collision rate and the distributions of minimum TTC and PET per run.
        The Wilcoxon rank-sum test was applied to the TTC and PET distributions with a significance level of 0.05.
        Two timeout cases occurred under the Proposed condition; runs with recorded TTC values were included in the TTC analysis, and runs in which the ego vehicle did not traverse the intersection were excluded from the PET calculation (three Proposed runs excluded).

\begin{figure}[t]
            \centering
            \includegraphics[width=\linewidth]{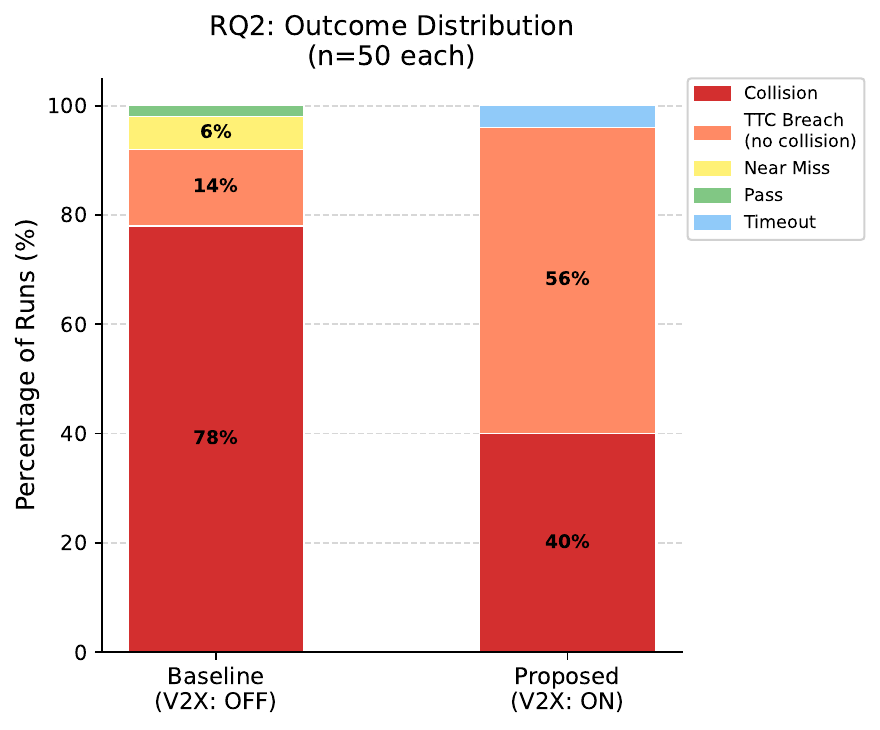}
            \caption{Outcome distribution for RQ2. V2X information reduces collisions from 78\% to 40\%, with an increase in TTC breach (no collision) cases}
            \label{fig:rq2_outcome}
        \end{figure}

        V2X information substantially reduced the collision rate.
        As shown in Figure~\ref{fig:rq2_outcome}, the collision rate under the Baseline condition was 78.0\% (39 out of 50 runs), whereas the collision rate under the Proposed condition decreased to 40.0\% (20 out of 50 runs).
        The non-collision rate improved from 22.0\% to 60.0\%, representing an approximately 2.7-fold improvement.
        TTC threshold violations without collision increased from seven to 28 cases, showing that many runs remained safety-critical despite not resulting in a collision.

\begin{figure}[t]
            \centering
            \includegraphics[width=\linewidth]{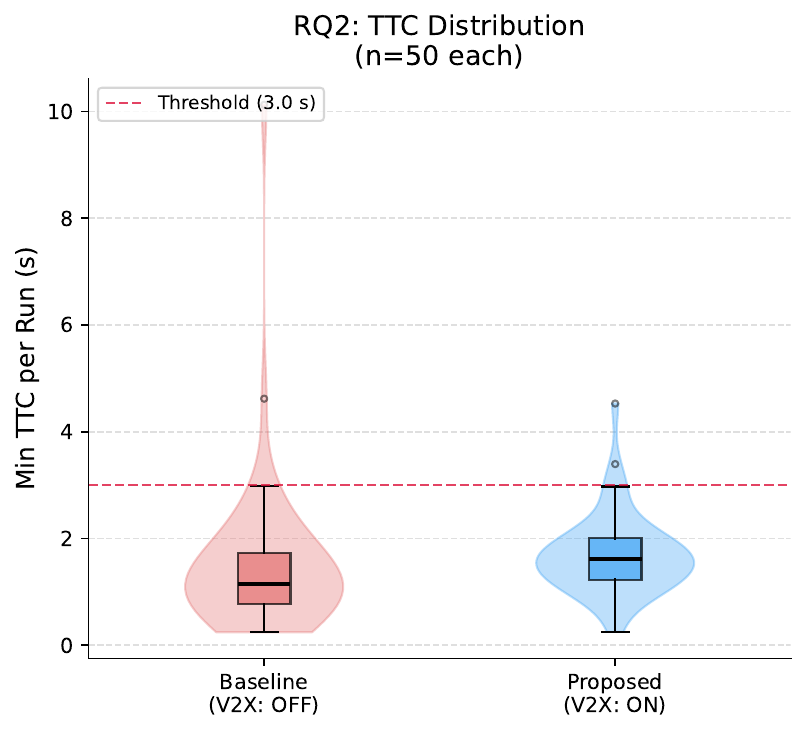}
            \caption{Distribution of minimum TTC per run in RQ2. The dashed line indicates the TTC threshold of 3.0\,s for edge-case detection.}
            \label{fig:rq2_ttc}
            \vspace{-4mm}
        \end{figure}

        Statistically significant improvement was also observed in the TTC distribution.
        As shown in Figure~\ref{fig:rq2_ttc}, the median of the minimum TTC per run improved from 1.155\,s under the Baseline condition to 1.610\,s under the Proposed condition.
        The standard deviation decreased from 1.484\,s to 0.743\,s, and cases with TTC $<$ 1.0\,s decreased from 32.0\% (16 out of 50 runs) to 10.2\% (five out of 49 runs).
        The Wilcoxon rank-sum test confirmed a statistically significant TTC improvement ($p = 0.0071$).

\begin{figure}[t]
            \centering
            \includegraphics[width=\linewidth]{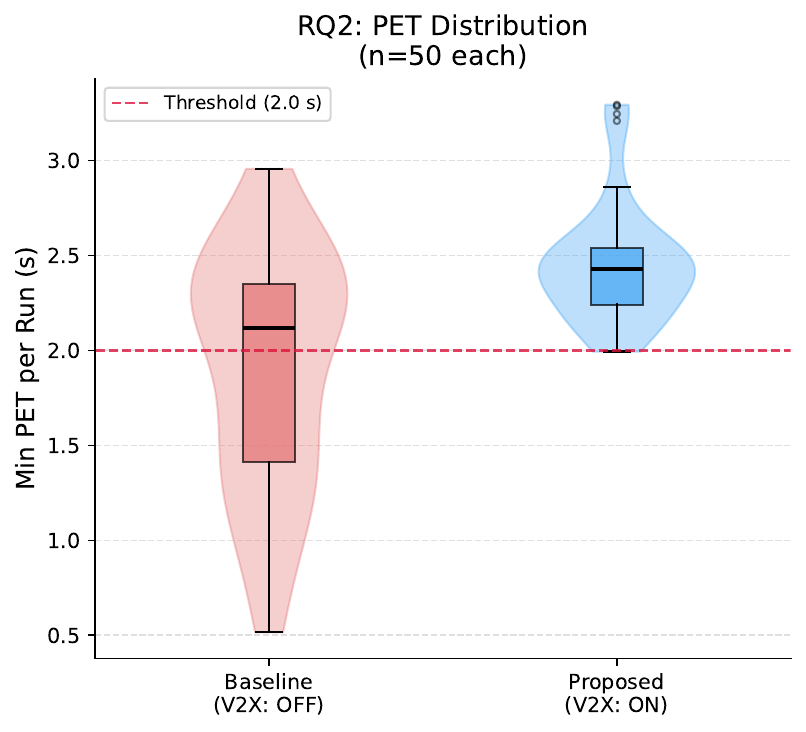}
            \caption{Distribution of minimum PET per run in RQ2. The dashed line indicates the PET threshold of 2.0\,s for edge-case detection.}
            \label{fig:rq2_pet}
            \vspace{-4mm}
        \end{figure}

        The PET distribution showed a stronger shift.
        As shown in Figure~\ref{fig:rq2_pet}, the median of the minimum PET per run improved from 2.116\,s under the Baseline condition to 2.428\,s under the Proposed condition.
        Cases with PET $<$ 2.0\,s decreased sharply from 49.0\% (17 out of 35 runs) under the Baseline condition to 2.1\% (one out of 47 runs) under the Proposed condition.
        The minimum PET under the Proposed condition was 1.994\,s, and the PET distribution difference was highly significant ($p = 0.0001$).
        Because PET measures the time gap after intersection traversal, this result indicates that V2X information affected Autoware path planning and increased the crossing-time margin.

        The improvement results from two effects.
        First, SIM-LDM fuses the position and velocity of occluded NPC vehicles and pedestrians into Autoware perception, enabling earlier deceleration.
        Second, Dynamic Map information helps Autoware estimate intersection occupancy and shift crossing timing.
        The two Proposed timeout cases indicate a trade-off: additional information can also induce conservative behavior that prevents goal arrival within the time limit.

        \begin{figure}[t]
            \centering
            \includegraphics[width=\linewidth]{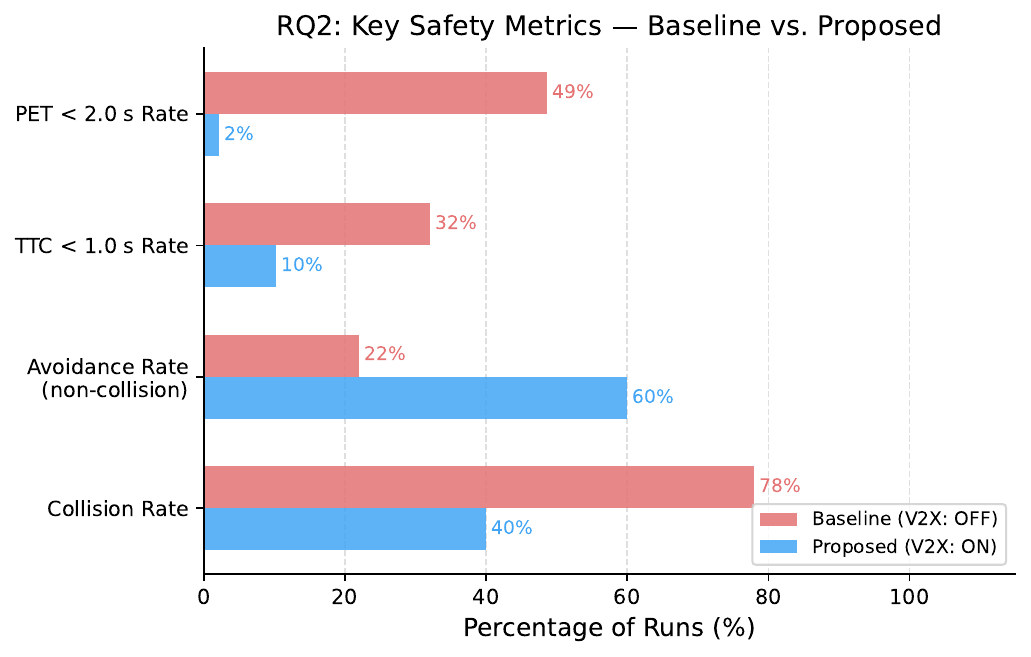}
            \caption{Collision rate comparison in RQ2. V2X information reduces the collision rate by 38 percentage points.}
            \label{fig:rq2_collision_rate}
            \vspace{-6mm}
        \end{figure}

        Remaining collisions mainly stemmed from the NPC behavior model.
        NPCs did not yield to vehicles on their route, and NPCs approaching from both sides could enter the intersection in sampling-dependent orders.
        In several cases, the ego vehicle decelerated for a far NPC detected via V2X, after which a near non-yielding NPC entered laterally and caused a collision.
        These cases do not negate the effect of V2X information; instead, they show that richer information can interact with non-yielding NPC behavior.

        These results answer RQ2: V2X information via Dynamic Map reduced the collision rate from 78.0\% to 40.0\% and improved the non-collision rate from 22.0\% to 60.0\%.
        Statistically significant TTC and PET improvements (TTC: $p = 0.0071$, PET: $p = 0.0001$) demonstrate that Dynamic Map information improves Autoware behavior in occluded intersection scenarios.

    \subsection{RQ3: System Performance Scalability}\label{subsec:RQ3}
        RQ3 evaluates the effect of increasing scenario complexity on system performance.
        Complexity is defined as the number of NPC vehicles in a scenario, with five conditions set at one, two, four, eight, and 16 vehicles.
        Five scenarios were generated and executed under each condition.
        The measured items are scenario generation time, simulation wall time, real-time ratio, CPU utilization, GPU utilization, GPU memory consumption, and RAM consumption.

        Scenario generation time is defined as the elapsed time from the start of Scenic sampling to the completion of concrete scene generation.
        Simulation wall time is measured as the elapsed wall-clock time from the start to the end of each scenario run.
        The real-time ratio is defined as the logical simulation time divided by the wall time; larger values indicate faster-than-real-time execution.
        Resource utilization is recorded as CPU utilization, GPU utilization, GPU memory consumption, and RAM consumption during simulation execution.

        Scenario generation time remained nearly constant as the number of NPCs increased.
        As shown in Table~\ref{tab:rq3_scene_gen}, the mean generation time was 0.084\,s with one NPC and 0.093\,s with 16 NPCs, an increase of only 0.009\,s (approximately 10.7\%).
        All conditions fell within 0.082--0.093\,s, indicating no notable complexity dependence in scenario generation cost.

        \begin{table}[t]
            \centering
            \caption{Average scenario generation time, simulation wall time, and real-time ratio per NPC count in RQ3.}
            \vspace{-2mm}
            \label{tab:rq3_scene_gen}
            {\renewcommand{\arraystretch}{1.3}
            \begin{tabular}{|r|r|r|r|}
                \hline
                \textbf{NPC Count} & \textbf{Gen. Time (s)} & \textbf{Wall Time (s)} & \textbf{RT Ratio} \\ \hline
                1  & 0.084 & 111.0 & 0.380$\times$ \\
                2  & 0.082 & 102.8 & 0.351$\times$ \\
                4  & 0.087 & 138.6 & 0.310$\times$ \\
                8  & 0.089 & 153.1 & 0.252$\times$ \\
                16 & 0.093 & 155.4 & 0.197$\times$ \\ \hline
            \end{tabular}}
            \vspace{-4mm}
        \end{table}

\begin{figure}[t]
            \centering
            \includegraphics[width=\linewidth]{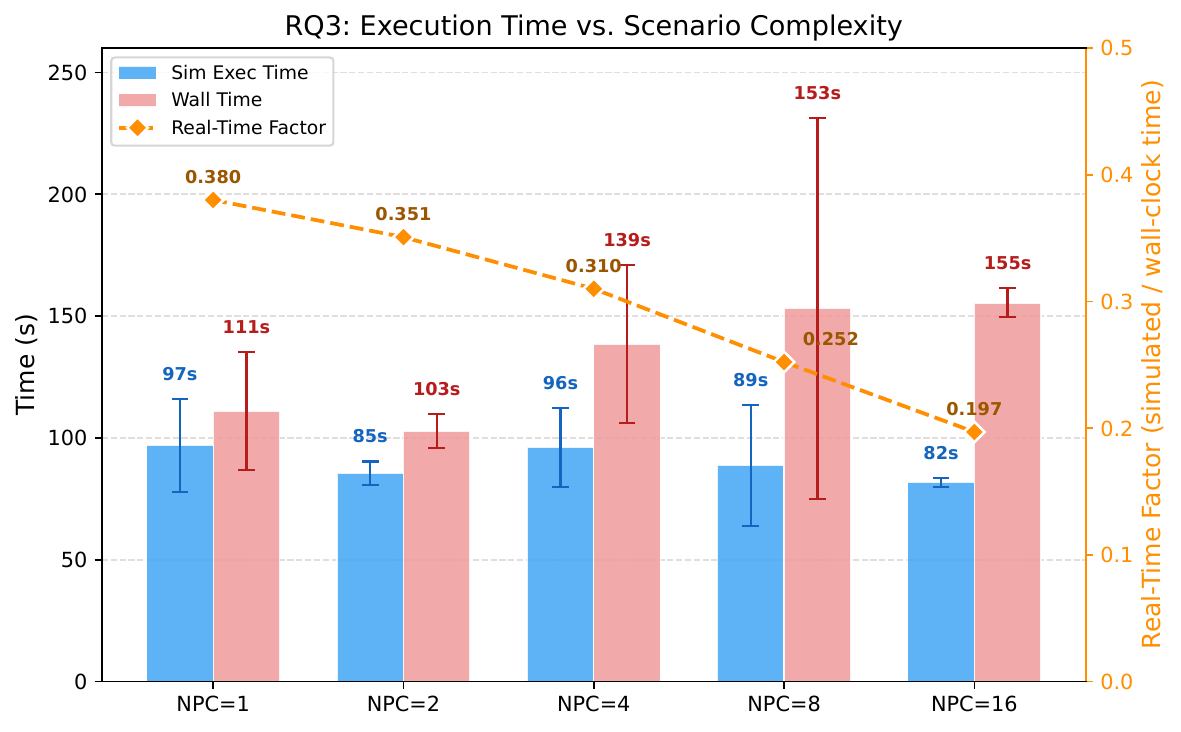}
            \caption{Wall time and real-time ratio per NPC count in RQ3.}
            \label{fig:rq3_execution_time}
            \vspace{-4mm}
        \end{figure}

        Simulation execution cost increased with NPC count.
        As shown in Table~\ref{tab:rq3_scene_gen}, wall time increased from 111.0\,s with one NPC to 155.4\,s with 16 NPCs, corresponding to an increase of approximately 40\%.
        The wall time and real-time ratio per NPC count are shown in Figure~\ref{fig:rq3_execution_time}.

\begin{figure}[t]
            \centering
            \includegraphics[width=\linewidth]{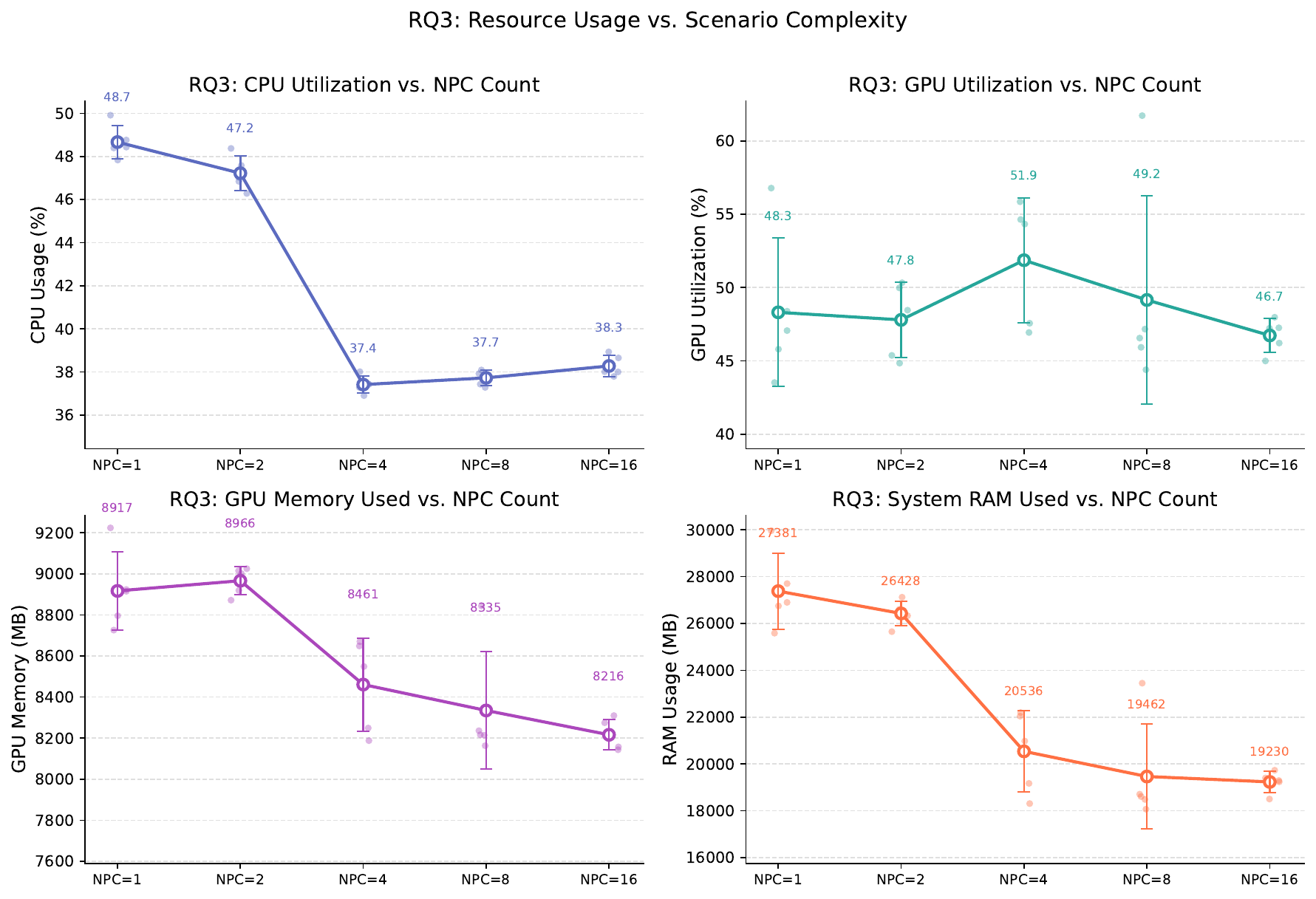}
            \caption{Resource usage for up to 16 NPCs in RQ3.}
            \label{fig:rq3_resource_usage}
            \vspace{-6mm}
        \end{figure}

        Resource utilization did not exhibit monotonic degradation with increasing NPC count.
        As shown in Figure~\ref{fig:rq3_resource_usage}, CPU utilization was approximately 47--49\% with one to two NPCs but converged to approximately 37--38\% with four or more NPCs.
        GPU utilization remained within approximately 45--52\%, and GPU memory consumption fluctuated within approximately 8,200--8,900\,MB.
        RAM consumption averaged 27,381\,MB with one NPC, whereas it was approximately 19,000--20,500\,MB with four or more NPCs.
        This non-monotonic behavior suggests that computational load also depends on traffic interaction density and scenario termination timing.

        These results answer RQ3 by showing that Scenic sampling is not the throughput bottleneck.
        Scenario generation remained below 0.1\,s for up to 16 NPCs, whereas simulation wall time increased and the real-time ratio decreased as NPC count increased.
        CPU, GPU, GPU memory, and RAM usage remained bounded, and VAST executed integrated simulations continuously with up to 16 NPCs.
        In practice, the fixed restart cost of AWSIM and Autoware is a larger limitation than computational throughput because the current implementation restarts both systems between scenarios.
        A persistent simulation environment is therefore the main future direction for improving large-scale scenario exploration throughput.

\subsection{Lessons Learned}\label{subsec:lessons}


The evaluation provides three lessons for V2X/Dynamic Map-aware validation in Autoware/AWSIM environments.

First, the generation component contributes by translating Lanelet2 geometry and probabilistic specifications into reproducible edge-case executions, not by introducing a new search algorithm.
This indicates that, for toolchains targeting realistic autonomous-driving stacks, map compatibility and executable scenario translation are critical design requirements alongside the scenario sampling strategy.

Second, paired execution with and without Dynamic Map information exposes how infrastructure-derived object information changes Autoware behavior under the same generated scenario.
Because the compared runs are generated from the same scenario distribution, the observed behavioral differences can be attributed to V2X-derived Dynamic Map information rather than to uncontrolled scenario variation.

Third, practical throughput is limited more by AWSIM/Autoware execution and restart overhead than by Scenic sampling.
This suggests that future V2X/Dynamic Map-aware validation tools should prioritize persistent simulator execution, automated state reset, and middleware-level lifecycle management before optimizing the scenario sampler.

\section{Related Work}\label{sec:Related}
    Related research is reviewed from the perspective of V2X/Dynamic Map-aware validation.
    The discussion is organized into four categories: scenario description languages, autonomous driving simulators, scenario-based testing approaches, and Dynamic Maps.
    
    \subsection{Scenario Description Languages}
        Scenario description languages play an important role in testing autonomous driving systems.
        OpenSCENARIO~\cite{OpenSCENARIO_XML,OpenSCENARIO_DSL} is a representative scenario description language.
        As an international standard for describing dynamic scenarios, this language is widely used in virtual development, verification, and validation of ADAS and autonomous driving.
        Two representation formats exist: XML-based and DSL-based, with both formats being developed in parallel.
        The XML-based format, however, tends to result in verbose and complex descriptions, while the DSL-based format is still evolving and has a steep learning curve.
        
        Scenic~\cite{Scenic} is proposed as a probabilistic scenario description language to address these limitations of OpenSCENARIO.
        This language represents scenarios as probability distributions over scenes and enables diverse test case generation through sampling.
        Native support for simulators such as CARLA is provided; however, direct integration with the AWSIM and Autoware testing infrastructure is not currently supported.
        VAST addresses this limitation by extending SS2 to interpret Scenic programs and execute them on AWSIM together with Autoware.
        
    \subsection{Autonomous Driving Simulators}
        Autonomous driving simulators are required to execute scenarios and evaluate autonomous driving systems safely and efficiently.
        CARLA~\cite{CARLA} is an open-source urban driving simulator built on Unreal Engine and provides official Scenic integration for probabilistic scenario-based testing, but CARLA primarily targets the ROS~1 ecosystem and lacks native support for the ROS~2-based architecture of Autoware.
        AWSIM~\cite{AWSIM} is a Unity-based simulator optimized for Autoware, uses ROS~2-native communication, and supports OpenSCENARIO-based testing through SS2~\cite{SS}, but AWSIM does not currently support probabilistic scenario description languages such as Scenic.
        VAST bridges this gap by enabling Scenic-based scenario specification for AWSIM.
    
    \subsection{Scenario-based Testing}
        Effective testing on simulators requires the preparation of scenarios with sufficient quantity and diversity.
        Scenario-based testing requires a vast number of scenarios, with particular emphasis on edge cases.
        Automated generation methods have been researched because creating edge-case scenarios requires substantial effort.
        Examples include a method for automatically generating edge-case scenarios in OpenSCENARIO format from autonomous driving disengagement reports~\cite{SEAMS2024_Song} and a method for generating safety-critical scenarios using large language models~\cite{ChatScene}.
        These approaches enable efficient scenario generation; however, the number of generated scenarios is limited, making them insufficient for comprehensive edge-case exploration.
        
        Data-driven approaches that learn realistic traffic behavior from real-world datasets have also been proposed.
        TrafficGen~\cite{TrafficGen} employs an autoregressive generative model to synthesize diverse traffic scenarios from fragmented driving data.
        Adversarial scenario generation methods, which optimize scenarios to maximize collision probability, have demonstrated effectiveness in discovering edge cases.
        STRIVE~\cite{STRIVE} performs adversarial optimization in the latent space of a learned traffic model, while AdvDiffuser~\cite{AdvDiffuser} utilizes guided diffusion models for transferable adversarial scenario generation.
        Importance sampling approaches~\cite{RareEvent_NeurIPS} accelerate rare-event discovery by biasing the sampling distribution toward dangerous situations.
        These methods, however, primarily target single-vehicle testing environments and do not address V2X communication or Dynamic Map integration.

        Search-based testing has also been applied to autonomous-driving control systems.
        Learnable evolutionary testing~\cite{LearnableEA_ADAS} combines multiobjective evolutionary search with decision tree classification to generate critical scenarios for vision-based control systems and characterize critical regions of the input space.
        FITEST~\cite{FITEST} formulates feature-interaction testing for autonomous cars as a many-objective search problem and uses hybrid test objectives to expose failures caused by interactions among driving features.
        These approaches provide effective guidance for simulation-based fault discovery, whereas VAST focuses on reproducible V2X/Dynamic Map-aware validation by integrating probabilistic scenario specification, Dynamic Map injection, and Autoware/AWSIM execution.
        
        An alternative approach for scenario generation in V2X environments involves utilizing data obtained from real-world environments~\cite{Detection_CornerCase}.
        This method installs multiple cameras, LiDARs, weather radars, and V2X-enabled traffic signals at intersections for real-time data collection and analysis.
        Results report that 24 edge-case scenarios were identified and generated after approximately six months of data collection.
        The approach offers high realism because the generated scenarios are derived from real-world data; however, such data-driven scenario generation requires substantial effort and cost for data collection and analysis.
        
    \subsection{Dynamic Maps}
        Dynamic Maps play an important role in V2X environments, which are the target of this research.
        Various approaches for Dynamic Map research have been proposed.
        A graph-based LDM~\cite{iLDM} is proposed to address the scalability and flexibility limitations of relational databases.
        A framework for constructing Dynamic Maps in edge computing environments~\cite{liveMap} and an integrated platform for computation offloading and dynamic data sharing~\cite{offload} also exist.
        Edge Dynamic Map architecture~\cite{EDM_architecture_for_C-ITS} combines Multi-access Edge Computing with time-series databases and enhances C-ITS by leveraging MEC and 5G technologies.
        Existing Dynamic Map studies mainly address data management and edge architectures, whereas VAST evaluates how Dynamic Map information affects a ROS~2-based autonomous-driving stack by injecting the information into Autoware during simulation.
        First Mile~\cite{First_Mile} is an open experimental environment for V2X equipped with 89 sensors along a 3.5~km public road section, contributing to the reduction of the simulation-to-reality gap and providing real-world datasets.
        
    \subsection{Research Positioning}
        Based on the aforementioned related work, the positioning of this research is clarified.
        A comparison of the proposed framework with existing research is presented in Table~\ref{tab:related_work}.
        
        \begin{table}[t]
            \centering
            \caption{Comparison of the proposed framework with existing studies}
            \vspace{-2mm}
            \label{tab:related_work}
            {
                \renewcommand{\arraystretch}{0.9}
                \begin{tabularx}{\columnwidth}{|X|c|c|c|c|c|} \hline
                    ~                               & \textbf{OSS} & \textbf{ASG} & \textbf{PSG}  & \textbf{DMS} & \textbf{CEC} \\ \hline
                    OpenSCENARIO~\cite{OpenSCENARIO_XML}    & \checkmark   & ~            & ~            & ~            & ~            \\ \hline
                    AWSIM + SS~\cite{AWSIM,SS}              & \checkmark   & ~            & ~            & ~            & ~            \\ \hline
                    Scenic + CARLA~\cite{Scenic,CARLA}      & \checkmark   & \checkmark   & \checkmark   & ~            & \checkmark   \\ \hline
                    TrafficGen~\cite{TrafficGen}            & \checkmark   & \checkmark   & \checkmark   & ~            & ~            \\ \hline
                    STRIVE~\cite{STRIVE}                    & ~            & \checkmark   & \checkmark   & ~            & \checkmark   \\ \hline
                    AdvDiffuser~\cite{AdvDiffuser}          & ~            & \checkmark   & \checkmark   & ~            & \checkmark   \\ \hline
                    RareEvent~\cite{RareEvent_NeurIPS}      & ~            & ~            & \checkmark   & ~            & ~            \\ \hline
                    Learnable EA~\cite{LearnableEA_ADAS}    & ~            & \checkmark   & ~            & ~            & \checkmark   \\ \hline
                    FITEST~\cite{FITEST}                    & ~            & \checkmark   & ~            & ~            & \checkmark   \\ \hline
                    ChatScene~\cite{ChatScene}              & ~            & \checkmark   & \checkmark   & ~            & ~            \\ \hline
                    iLDM~\cite{iLDM}                        & ~            & ~            & ~            & \checkmark   & ~            \\ \hline
                    liveMap~\cite{liveMap}                  & ~            & ~            & ~            & \checkmark   & ~            \\ \hline
                    EDRP and LDMP~\cite{offload}            & ~            & ~            & ~            & \checkmark   & ~            \\ \hline
                    EDM~\cite{EDM_architecture_for_C-ITS}   & ~            & ~            & ~            & \checkmark   & ~            \\ \hline
                    First Mile~\cite{First_Mile}            & ~            & ~            & ~            & \checkmark   & ~            \\ \hline
                    CornerCase~\cite{Detection_CornerCase}  & ~            & ~            & ~            & \checkmark   & \checkmark   \\ \hline
                    NLP-Scenario~\cite{SEAMS2024_Song}      & ~            & \checkmark   & ~            & ~            & \checkmark   \\ \hline
                    Proposed Framework                      & \checkmark   & \checkmark   & \checkmark   & \checkmark   & \checkmark   \\ \hline
                \end{tabularx}
            }
            \begin{minipage}{\columnwidth}
                \vspace{1mm}
                \raggedright
                \scriptsize
                \renewcommand{\arraystretch}{0.90}
                \begin{tabularx}{\columnwidth}{@{}l@{\hspace{1mm}}>{\raggedright\arraybackslash}X@{\hspace{3mm}}l@{\hspace{1mm}}>{\raggedright\arraybackslash}X@{}}
                    OSS: & Open-Source Software &
                    ASG: & Automatic Scenario Generation \\
                    PSG: & Probabilistic Scenario Generation &
                    DMS: & Dynamic Map Support \\
                    \multicolumn{4}{@{}l@{}}{CEC: Constraint-based Edge-Case Generation}
                \end{tabularx}
            \end{minipage}
            \vspace{-6mm}
        \end{table}
        
        In contrast to these related works, VAST provides an integrated validation toolchain that can execute controllable stress scenarios, inject Dynamic Map information, and compare Autoware behavior with and without V2X-derived inputs in a reproducible simulation environment.
        Scenic is Python-based, offering superior readability, maintainability, and ease of learning compared to OpenSCENARIO DSL or XML.
        The main contribution of this research is not scenario generation alone, but the integration of probabilistic scenario specification, Lanelet2-compatible execution, Dynamic Map data delivery, paired V2X evaluation, and throughput measurement in an Autoware/AWSIM toolchain.

\section{Conclusion}\label{sec:Conclusion}
This paper presented VAST, a V2X/Dynamic Map-aware validation toolchain that integrates Scenic, Scenario Simulator v2, AWSIM, Autoware, and SIM-LDM for system-level evaluation of cooperative autonomous driving. Rather than proposing a new scenario-search algorithm, VAST addresses the interoperability challenge of executing Lanelet2-compatible probabilistic edge case scenarios in an Autoware/AWSIM environment and evaluating their effects through a Dynamic Map data path.

For RQ1, the Lanelet2-compatible constrained generation pipeline increased the edge case discovery rate from 40.0\% to 80.0\% and reduced the average time per discovered edge case from 259.7\,s to 110.4\,s. For RQ2, enabling Dynamic Map information under the same generated scenario distribution changed the system-level behavior of Autoware, reducing the collision rate from 78.0\% to 40.0\% and increasing non-collision outcomes from 22.0\% to 60.0\%, with statistically significant TTC/PET shifts. 
For RQ3, scenario sampling remained below 0.1\,s with 16 NPCs, while AWSIM/Autoware execution and restart overhead dominated runtime; the real-time factor decreased from 0.380 to 0.197 as complexity increased.

These results show that VAST provides a practical validation infrastructure for V2X/Dynamic Map-aware cooperative autonomous driving and a concrete toolchain instance for evaluating distributed cyber-physical behavior across the IoT-to-Edge-to-Cloud continuum.
To support reproducibility and further development, the core components of the proposed toolchain are publicly available at \url{https://github.com/azu-lab/VAST}.
Future work will evaluate degraded cooperative-perception conditions by modeling delay, packet loss, jitter, and stale Dynamic Map updates. 
Additional extensions include uncertainty-aware V2X injection, interactive NPC behavior, optimization-based scenario search, and persistent AWSIM/Autoware execution.

\section*{Acknowledgment}
This work was supported by JST FOREST Grant Number JPMJFR242G and JSPS KAKENHI Grant Number 23H00464.

\bibliographystyle{ieeetr}    
\bibliography{ref}

\end{document}